\documentclass{article}
\usepackage{ijcai21}

\usepackage{times}

\usepackage{soul}
\usepackage{url}
\usepackage[hidelinks]{hyperref}
\usepackage[utf8]{inputenc}
\usepackage[small]{caption}
\usepackage{graphicx}
\usepackage{amsmath}
\usepackage{booktabs}
\usepackage{multirow}
\usepackage{amsfonts}
\usepackage{graphics}
\usepackage{algorithm}
\usepackage{algorithmic}

\usepackage{xcolor}

\newcommand{\tocheck}[1]{\textcolor{black}{#1}}

\title{MinMaxCAM: Improving object coverage for CAM-based \\Weakly Supervised Object Localization}
\author{
Kaili Wang$^{1,2}$\and
Jose Oramasr$^2$\and
Tinne Tuytelaars$^{1}$\\
$^1$KU Leuven, ESAT-PSI\\
$^2$University of Antwerp, imec-IDLab\\
}

\begin{document}

\maketitle

\begin{abstract}
One of the most common problems of weakly supervised object localization is that of inaccurate object coverage. In the context of state-of-the-art methods based on Class Activation Mapping, this is caused either by localization maps which focus, exclusively, on the most discriminative region of the objects of interest, or by activations occurring in background regions.
%
% Interestingly, an analysis on the localization maps generated by different backbones reveals that different backbones actually suffer from different problems.
% Opposite from the well-known limited coverage issue, we found that the localization map generated by frequently-used backbones (e.g. ResNet, MobileNet) activate highly on background regions.
%
% For some frequently-used backbones whose output of the last convolutional block is too small (i.e. $7\times 7$), a common solution is to increase its size by changing the stride in some convolution blocks.
% The localization map generated by them (e.g. ResNet, MobileNet), however, activates highly on background regions.
%
To address these two problems, we propose two representation regularization mechanisms: \textit{Full Region Regularization} which tries to maximize the coverage of the localization map inside the object region, and \textit{Common Region Regularization} which minimizes the activations occurring in background regions.
%
% \textit{Full Region Regularization (FRR)} and \textit{Common region Regularization (CRR)}.
% \textit{FRR} tries to maximize the coverage of the localization map while \textit{CRR} minimizes it when the localization map activates highly in the background.
%
We evaluate the two regularizations on the ImageNet, CUB-200-2011 and OpenImages-segmentation datasets, 
% for the VGG16, ResNet50 and MobilenetV2 backbones. 
% Our extensive experiments 
and show that the proposed regularizations tackle both problems, outperforming the state-of-the-art by a significant margin.
% achieving the new state

% One of the most used methods for weakly supervised object localization is based on Class Activation Map, where the size of the output of the last convolutional block can influence the quality of the localization map.

\end{abstract}

\section{Introduction}

Learning how to localize objects in images without relying on data paired with expensive location-specific annotations is a highly desirable capability. Therefore, it is no surprise that this task, usually referred to as Weakly-supervised Object Localization  (WSOL), has gained attention in recent years. 

One of the most used methods for this task is based on Class Activation Map (CAM) \cite{zhou2016learning},
see
\cite{singh2017hide,zhang2018self,zhang2018adversarial,ADL2019,yun2019cutmix,zhang2020inter}.
In these works, it
has been noticed that the localization map generated by CAM
%-based methods 
focuses on the most discriminative region of the image.
The reason is simple: the backbone is trained for classification since there is no access to the coordinates of the object, so it learns the discriminative features for each class. As a result, the object coverage is under-estimated.

%%% Existing methods
%%  - perturbation-based methods -> spatial dropout
%%  - updating convolutional layers vs learning how to re-weight them
%%  - need additional networks

Existing efforts to address this problem follow one of three common strategies.
They either iteratively occlude/replace relevant regions of input in order to force the model to learn features that enable localization~\cite{singh2017hide,yun2019cutmix},  or rely on additional networks to assist with the localization task~\cite{zhang2018self,zhang2018adversarial}. 
% It is clear that these two strategies increase the complexity of the model.
%
Alternatively, a more simple strategy is to update the representation learned by the convolutional layers of the model~\cite{ADL2019,zhang2020inter}. 
% However, this comes at the cost of increased complexity during  training.

\begin{figure}
\centering
\includegraphics[width=0.5\textwidth]{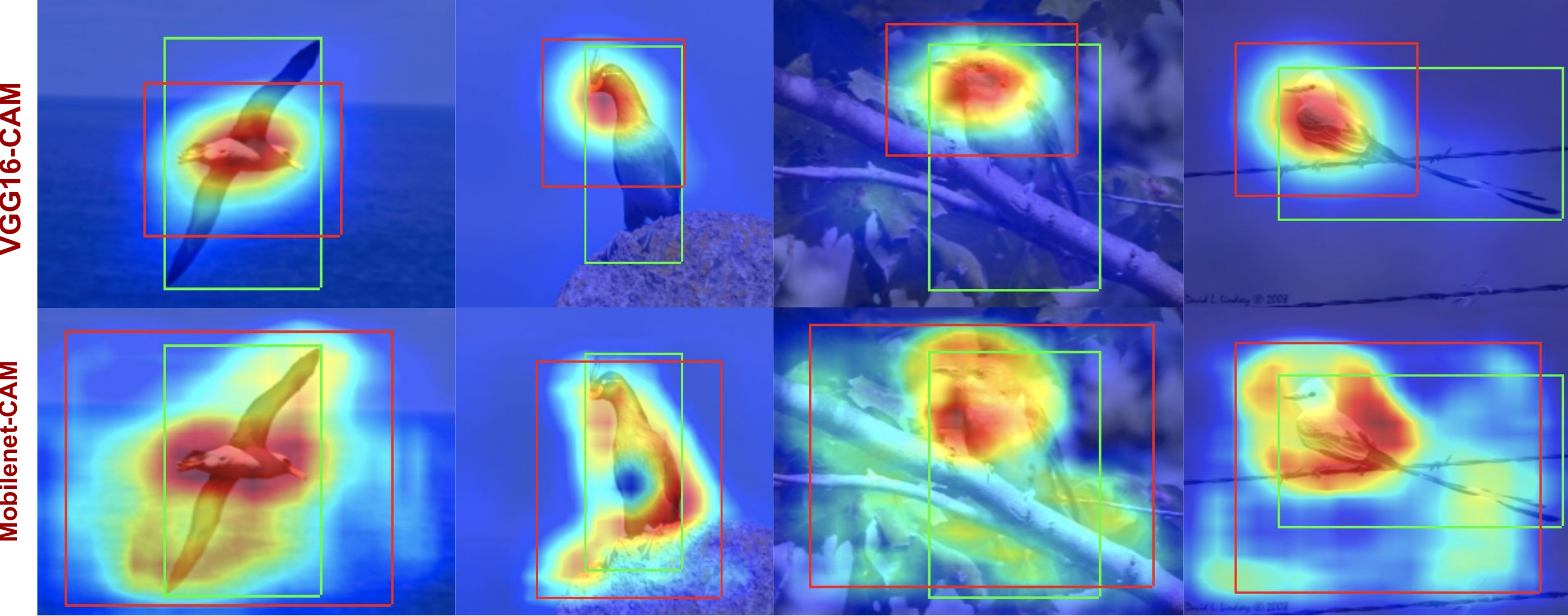}
\caption{CAM localization maps generated with the VGG16 (top) and MobileNet (bottom) backbones with their estimated bounding boxes (red) and ground-truth (green).
% We present the ground truth (green) while the predicted bounding box (red) estimated by the localization map. 
}
\label{fig:shortage}
% \vspace{-4mm}
\end{figure}

For CAM-based methods, the localization map is generated by a weighted linear combination of the feature map of the last convolutional layer and then rescaled to the image's size.
If the size of the feature map is too small, say $7\times 7$ while the input size is $224 \times 224$, the resized localization map will have poor precision.
In order to increase the size while still loading the original pre-trained weights of the backbone,
a common strategy is to change the stride in some convolutional block
\cite{ADL2019,zhang2020inter,choe2020wsol}.
However, we observe the localization map generated this way can activate a lot on the background, i.e.~the object coverage is over-estimated (see Fig.~\ref{fig:shortage}).
To solve it, activations on the background  should be suppressed.
% Besides, 
To this end, \cite{CCAM_20} 
%also noticed this issue and 
%address this by 
%suggested to 
compute all possible CAM of one image and use some pre-defined combination functions to combine them.
Clearly, this method costs more computation and the combination functions are not optimized.

Then the research question for us is to find a way to actively control the activation distribution on the localization map, maximizing or minimizing the activations
% accordingly.
as needed.
%
%To address these weaknesses, 
% We begin from the observation that a classification model usually learns relevant features that are located either on instances of the classes of interest, or in background regions with high co-occurrence with these classes. 
We propose \textit{MinMaxCAM}, a method that learns how to re-weight the feature maps initially learned by the classification model so that it is capable of shifting the mass of their internal activations. This not only enables accurate object localization but is relatively stable to train and does not   need additional networks.
% In the end,
In particular,
we design two regularizations, \textit{Common Region Regularization (CRR)} and \textit{Full Region Regularization (FRR)}, that can serve as objective functions for the model to optimize the linear layer after global average pooling (GAP).
\textit{CRR} encourages
multiple images from the same class to share similar foreground representations:
% Therefore, by minimizing the distance of representations from 
% In training, we obtain the localization map and 
the representations extracted from the common foreground regions should be close to each other.
During training,
the generated localization map is used to extract the localized-object representation.
By minimizing the distance between these representations,
the model can optimize the localization map.
% which means minimizing the activations on the localization map.
On the other hand, inner-class differences can reduce the common region to a part of the object only. The same may happen due to failed localization of the most discriminative region. To tackle these situations, \textit{FRR} is proposed. It  
% maximizes 
stimulates covering a larger part of the object.
% the activations on the localization map to cover a larger area.

% Our proposed method 
\textit{MinMaxCAM}
has a number of advantages: 
i) It is light-weight: it only relies on a standard classification model; no extra network is needed. It saves computation resources and is relatively simple to train.
ii) 
% our proposed method not only enlarges the localization map to cover more region of the foreground object,  but also suppresses the activations that are highly activated in background regions.
%, for some cases where the localization map is activated a lot on in the background region.
The proposed method produces more precise or tighter bounding boxes.
iii) Despite its simplicity, the proposed method is capable of setting a new state-of-the-art performance on the ImageNet,  CUB-200-2011 and OpenImages-segmentation datasets, outperforming existing methods by a significant margin.

% 1. None-parameter method, save resources, easy to train, only two hyper-parameters. No extra network is needed. (When show performance, show the number para.)
% 2. Can be plugged in to other systems easily
% 3. New direction of object localisation 
% 4. Not only enlarge the CAM, but also make CAM more precise, -> both enlarge and make it smaller
%     1. Adjust the localisation map, according to the architecture, previous method can only enlarge the feature map, while for some cases, like change the stride of resent or mobile net, the localisation map can increase, including the background  (check the score distribution over threshold)
% 5. New SOTA

%%============================================
%% RELATED WORK
%%============================================

\section{Related Work}
\label{sec:relatedwork}
% \subsection{Weakly Supervised Object Localization}
Most existing works related to WSOL~\cite{singh2017hide,zhang2018adversarial,zhang2018self,ADL2019,yun2019cutmix,zhang2020inter,CCAM_20} are based on Class Activation Mapping (CAM)~\cite{zhou2016learning}.
% In general, they estimate localization maps by using the feature maps of a neural network which is trained for a classification task.
%
They address the WSOL task, indirectly, by solving the problem that the generated localization map only focuses on the most discriminative regions of the image.
These methods can be divided into two types: non-parametric (w.r.t. CAM) and parametric methods.

%Among them, 
\cite{singh2017hide,ADL2019,yun2019cutmix,zhang2020inter} and our work belong to the first type.
These methods do not need extra networks during the training and inference phases.
% compared to \cite{zhou2016learning}, which makes the method lightweight, easy to implement and save computation resources.
This makes the methods lightweight, easy to implement and saves computation resources.
\cite{singh2017hide} force the neural networks to focus on other relevant regions of the objects of interest by randomly occluding some patches of the input image when the network is trained for a classification task.
\cite{yun2019cutmix} extend this idea by using patches from other images as occluding regions in a given image.
\cite{ADL2019} propose a simple but effective method: randomly drop out the most highly activated region or apply an attention mask on the feature maps when the classification network is trained.
\cite{zhang2020inter} use information shared by two images from the same class to improve the localization map. 
They apply two constraints to improve the quality of the localization map.
The first constraint is to learn the
% stochastic feature consistency
consistent features of two images of the same class 
by randomly sampling the features located in the most activated region and minimizing their distance. 
The second constraint is to compensate the limitation that features can only keep consistency within batches, where it 
learns a global class center for each class.
% By doing this, the model can learn to suppress activation on the different regions (e.g. background).
% The two constraints are similar to one of our proposed regularizations, i.e. \textit{CRR}.
% The constraints updates the parameters of the CNN. 
% They proposed to use two images from the same class to apply the constraints.
% Different from \cite{zhang2020inter}, our method uses the generated localization map to mask the original image and extract the feature in the training phase.
% Instead of updating the convolutional layers, we focus on the final prediction layer, which provides the weight for the generation of the localization map.
% In addition, we also consider the case where more than two images from the same class are available. 
\cite{CCAM_20} found that localization maps generated by CAM can also activate on the background. To suppress the activations of an image, they propose to compute all the possible CAMs firstly, and  combine them via a combination function.
This combination function is not learned during the training but pre-defined and related to the prediction probability of each possible class. Differently, our method only computes the localization map once for each image.

% In addition, our method also 

\cite{zhang2018adversarial,zhang2018self}
% modified the network architecture.
add extra components based on the CAM model.
\cite{zhang2018adversarial} proposes a two-head architecture where the activation map generated by one stream is used to suppress the most discriminative region of the activation map generated by another one. By doing this, the model learns to use information from other relevant regions instead of the most discriminative one.
\cite{zhang2018self} proposes to generate self-produced guidance (SPG) masks that separate the target object from the background. The masks are learned by the high-confidence regions within attention maps progressively and they also provide the pixel-level supervisory signal for the classification networks.

% \tocheck{ Regarding to the evaluation for the WSOL task, \cite{choe2020wsol,wsol_eval_journal_submission} proposed a new metric MaxBoxAcc and its revised version MaxBoxAccV2 to evaluate the estimated bounding box. It does not depend on a pre-defined threshold to binarize the localization map, nor the classification ability of the localization model.}

Different from these methods, we focus on the linear combination part of the CAM method rather than the feature extraction part,
or the structure of the input images. 
The linear layer is optimized based on the proposed regularizations, which provides the optimal combination factors to generate the localization map.
Similar to the non-parametric methods, our method does not have more trainable parameters. 
%In addition, w
We only introduce two more hyper-parameters which are the weights for the two regularizations.

% In addition, while
% \cite{singh2017hide,zhang2018adversarial,zhang2018self,ADL2019,yun2019cutmix} 
% focused on enlarging the localization map aiming at covering the whole object, \cite{zhang2020inter} tried to suppress the activations on the non-relevant region. 
% Our method can address both problems by adjusting the weights of the two proposed regularizations.   

% \subsection{Object Co-localization}
% Using information from a set of images from the same class to do object localization task is called co-localization.

%%============================================
%% METHODOLOGY
%%============================================
%%%%%%%%%-----------------------------------------------
%%%%%%%%%-----------------------------------------------
% %\vspace{-4mm}

%%%%%%%%%-----------------------------------------------
%%%%%%%%%-----------------------------------------------
\section{Methodology}
\label{sec:method}

\subsection{Problem statement}
\label{sec:problemstatement}
Class Activation Map (CAM)~\cite{zhou2016learning} is widely used 
to localize an object of interest in an image, in a weakly supervised 
manner.
Given a backbone $B$ applied to input image $I$, followed by GAP and a linear layer $F$, in which $w\in\mathbb{R}^{C \times K}$ is the weight matrix, where $C$ is the number of classes.
The CAM of an image is computed as:
{
%\small
\begin{align}
    CAM_{raw} = \sum_{k=0}^{K-1} w_{k}^{c}B(I)
    \label{eq:cam}
\end{align}
}%
%\vspace{-6mm}
{
%\small
\begin{align}
    CAM = \frac{CAM_{raw} - min(CAM_{raw})}{max(CAM_{raw}) - min(CAM_{raw})}
    \label{eq:camNorm}
\end{align}
}%
where $c$ is the class of image $I$. 
In short, the localization map (CAM) is a linear combination of the feature maps of the last convolutional block of the backbone $B$. 
The weights of this combinations are taken directly from the weights of the linear layer w.r.t.~the class which $I$ belongs to.
% The weight is the weight of the last linear layer w.r.t. the class
%
\cite{choe2020wsol,zhou2016learning} noted that localization maps generated by VGG often focus on the most discriminative region of the image, rather than on the whole object. 
We also observe that for different backbones $B$, the localization map can sometimes even cover the whole image, i.e. fail to localize the object. Fig. \ref{fig:shortage} shows some examples of the two cases. 
The second case has not been addressed in the literature yet.
Here we propose two regularizations to address these problems.

\subsection{Common region regularization}
\label{sec:CRR}
The idea of the \textit{Common region regularization (CRR)} is that different images depicting foreground objects of the same class should share similar features.
The localized-object features can be obtained by applying the localization map $H$ which is generated via Eq.~\ref{eq:cam}\&\ref{eq:camNorm} on the image $I$ and extract the feature of $I{\odot} H$, denoted as $f$. 
Therefore, $f {=} B(I{\odot} H)$, where $\odot$ refers to the element-wise multiplication.
To save computation resources, we use the same backbone $B$. 

Given $S$ different images from the same class, we have
{%%\small
\begin{align}
    CRR = \frac{1}{S(S-1)} \sum_{i=0}^{S-1}\sum_{j=0}^{S-1} ||f_{i} - f_{j}||_{2}^{2}
    \label{eq:CRR}
\end{align}
}%
\textit{CRR} calculates the pair-wise distance of the feature of $I{\odot} H$.  
The goal of \textit{CRR} is to localize the common region of a set of images from the same class. 
By minimizing it, it can \textbf{minimize} the activations on the localization map, i.e. suppress the activations in non-object regions of the images.

%To make \textit{CRR} work, there are two assumptions:
There are two conditions in place for CRR to work: 
1) a set of images from the same class should have different background and 2) $f$ should have different activation values for different backgrounds. The first assumption is dependent on the dataset. We will discuss the second assumption later.

\subsection{Full region regularization}
\label{sec:FRR}
The goal of \textit{CRR} is to make the localization map small in order to only focus on common regions of a set of images. However, this can have the side-effect of making the backbone only focus on a small part of the object if the objects have some inner difference in different images.
In addition, for the case of the failed localization on the most discriminative region,
% To compensate for this effect, 
we propose \textit{Full region regularization (FRR)} to enlarge the localization map.
{%%\small
\begin{align}
    FRR = \frac{1}{S} \sum_{i=0}^{S-1} ||f_{i} - f_{i}^{o}||_{2}^{2}
    \label{eq:FRR}
\end{align}
}%
$f^{o}$ is the feature of the original image $I$, i.e. $f^{o}{=}B(I)$ .
$FRR$ calculates the distance of the feature between the $I{\odot} H$ and $I$. Minimizing \textit{FRR} has the effect of \textbf{maximizing} the activations on the localization map.
Similar to \textit{CRR}, there is one assumption in place for \textit{FRR} to work: $B$ should not be invariant to changes in intensity. We will discuss it in Sec.\ref{sec:intensitychange}.

\subsection{Learning process}
\begin{figure}
\centering
\includegraphics[width=0.5\textwidth]{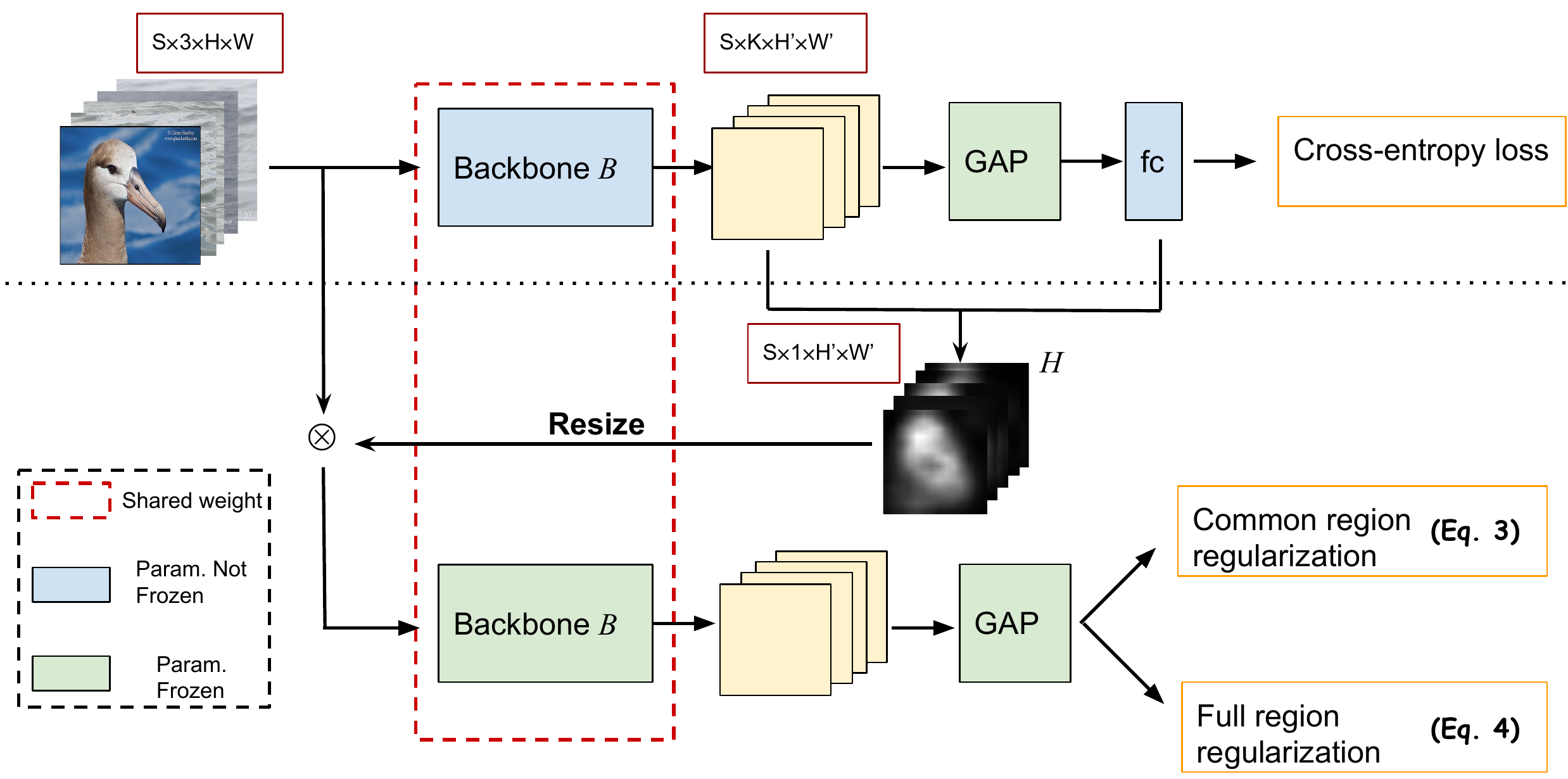}
\caption{Proposed method. In Stage I (above the dashed line) we train $B$ and the linear layer (fc) for a classification task. In Stage II (below) we multiply the localization map $H$ with the images and extract the representations using $B$ with its weights frozen. The representations are used to compute \textit{CRR} and \textit{FRR}. The two regularizations update the weights of the linear layer.
}
\label{fig:overallmodel}
% \vspace{-4mm}
\end{figure}

The learning process has two stages. 
% They update the model every batch. 
For stage I, it takes $N{\times} S$ images as input. The model is trained for the classification task, i.e. update the backbone and linear layer via the cross-entropy loss. It is the same as CAM.
{%\small
\begin{align}
    L_{S1} = -\sum_{i=1}^{N\times S} c_{i} log(\hat{y_i})
    \label{eq:CE}
\end{align}
}%
% where $c$ and $\hat{y}$ are the ground truth class and the predicted probability.
For stage II, the backbone $B$ is frozen as a feature extractor. 
It receives the images multiplied by the localization map ($I {\odot} H$) as input.
$H$ is generated via Eq.~\ref{eq:cam}\&\ref{eq:camNorm}.
This step introduces an intensity change on the original image. We discuss its effects in Sec.~\ref{sec:intensitychange}. 
The features ($f$ and $f^{o}$) extracted by $B$ are used for the two regularizations.
The loss only updates the weights of the linear layer, by minimizing $CRR$ and $FRR$.
{%%\small
\begin{align}
    L_{S2} = \lambda_1 CRR + \lambda_2 FRR
    \label{eq:CE}
\end{align}
}%
$L_{S1}$ and $L_{S2}$ update the model every mini-batch.
Fig.~\ref{fig:overallmodel} shows the proposed method.
To train our model there is no extra hyper-parameters besides the weights for the two regularizations.
During testing, the localization map is generated via CAM (Eq.~\ref{eq:cam}\&\ref{eq:camNorm}), therefore, there is no need for a set of images per class. 

% \begin{algorithm}[tb]
% \caption{Example algorithm}
% \label{alg:algorithm}
% \textbf{Input}: Your algorithm's input\\
% \textbf{Parameter}: Optional list of parameters\\
% \textbf{Output}: Your algorithm's output
% \begin{algorithmic}[1] %[1] enables line numbers
% \STATE Let $t=0$.
% \WHILE{condition}
% \STATE Do some action.
% \IF {conditional}
% \STATE Perform task A.
% \ELSE
% \STATE Perform task B.
% \ENDIF
% \ENDWHILE
% \STATE \textbf{return} solution
% \end{algorithmic}
% \end{algorithm}

%%%%%%%%%-----------------------------------------------
%%%%%%%%%-----------------------------------------------
%%%%%%%%%-----------------------------------------------
%%%%%%%%%-----------------------------------------------
\subsubsection{Why freeze the backbone ($B$) in Stage II?}
The goal of stage II is to adjust the weights $w_{k}^{c}$ which are used to generate the localization map by minimizing CRR and FRR.
% FRR provides the difference between masked feature $f$ and original feature $f^o$
$B$ serves as a feature extractor in this stage.
If $B$ were also updated, after minimizing \textit{FRR}, it would become invariant to intensity changes. In the later training process, \textit{FRR} could then not measure the difference between $f^{o}$ and~$f$.

%%=================================
%% EXPERIMENTS
%%=================================
%%%%%%%%%-----------------------------------------------
%%%%%%%%%-----------------------------------------------
\begin{figure*}
\centering
\includegraphics[width=1.\textwidth]{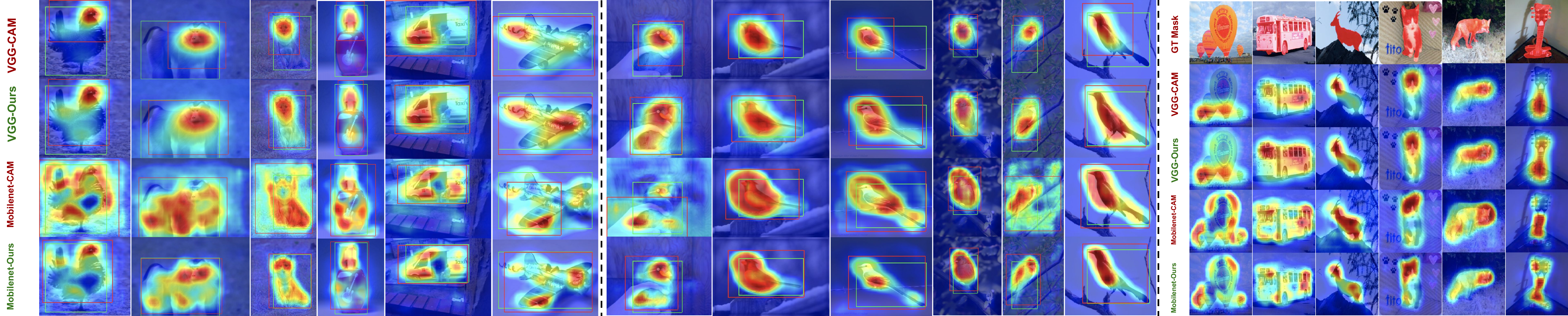}
\caption{Qualitative comparison of the localization map $H$ on the ImageNet, CUB and OpenImages dataset. 
For reference we show the ground truth bounding box (green) and the one estimated by $H$ (red) based on the optimal threshold $\tau$. %for different backbones and methods. 
For the OpenImages dataset, the first row shows the input image with the target segmentation mask.
}
\label{fig:qualityCUB}
% \vspace{-2mm}
\end{figure*}
%%%%%%%%%-----------------------------------------------
%%%%%%%%%-----------------------------------------------
\section{Experiments}
\label{sec:exp}
\subsection{Dataset and performance metric}
% We consider two widely used datasets CUB-200-2011 (CUB)~\cite{WahCUB_200_2011} and OpenImages instance segmentation subset~\cite{Openimages2019,choe2020wsol} to evaluate our method. CUB has 200 fine-grained classes of birds with 5,994 images for training and 5,794 images for testing. 
% %
% % Different from most methods
% % ~\cite{zhou2016learning,ADL2019,singh2017hide,yun2019cutmix,zhang2018adversarial,zhang2020inter}, 
% % % ~\cite{choe2020wsol}
% % where the models are tuned on the testing set due to the lack of a validation set in the original CUB-200-2011 dataset, 
% We follow \cite{choe2020wsol} to use the validation set to tune the model. The validation set contains 1,000 images in total with around 5 images per class. All the images from training, validation and testing set have the annotation of the bounding box of the bird.

% OpenImages instance segmentation subset (OpenImages)~\cite{wsol_eval_journal_submission} covers 100 classes. It contains 29,819, 2,500 and 5,000 images for training, validation and testing, respectively. Every image has the object segmentation as annotation.

%%===========================================================
We consider three widely used datasets ImageNet~\cite{imagenet_cvpr09}, CUB-200-2011 (CUB)~\cite{WahCUB_200_2011} and OpenImages instance segmentation subset~\cite{Openimages2019,choe2020wsol} to evaluate our method. 
ImageNet contains 1,000 classes with over 1 million images. 
% The training split is used to train the model while the validation split wi
Following \cite{choe2020wsol}, we use ImageNetV2~\cite{imagenetv2_2019} as the validation set to tune our model. 
This validation set contains 10 images per class with the object bounding boxes annotated by \cite{choe2020wsol}.
CUB has 200 fine-grained classes of birds with 5,994 images for training and 5,794 images for testing. 
Similarly, we follow \cite{choe2020wsol} to use a validation set collected by them to tune the model. The validation set contains 1,000 images in total, around 5 images per class. 
% All the images from training, validation and testing set have the annotation of the bounding box of the bird.
%
OpenImages instance segmentation subset (OpenImages)~\cite{wsol_eval_journal_submission} covers 100 classes. It contains 29,819, 2,500 and 5,000 images for training, validation and testing, respectively. Every image has the object segmentation as annotation.
%%===========================================================
\subsubsection{Performance metric}
\cite{zhou2016learning,ADL2019,yun2019cutmix,zhang2018adversarial,singh2017hide,zhang2020inter} use a pre-defined threshold (0.2) for the generated CAM to produce a localization region. \cite{choe2020wsol} argues that using a fixed pre-defined threshold can be disadvantageous for certain methods since the ideal threshold may depend on the data and architecture that are used. In short, for different datasets, architectures and methods, the ideal threshold is different. 
We follow this idea and use the metric proposed by \cite{choe2020wsol}.
For ImageNet and CUB datasets we use two threshold-free metrics to evaluate the localization map, i.e. MaxBoxAcc and MaxBoxAccV2.
MaxBoxAcc is equivalent to \textit{GT-known localization accuracy}
% ~\cite{} \textbf{REF},
where one localization map is counted correct when the intersection over union (IoU) of the estimation and ground truth bounding box is larger than 0.5.
Differently, to avoid using a fixed pre-defined threshold for binarizing the localization map, here we set various $\tau$ thresholds 
% (to binarize the localization map) 
to find the best performance.
MaxBoxAccV2 is the average of three MaxBoxAcc when the IoU is 0.3, 0.5 and 0.7. 
% For the case that multiple bounding boxes are in one image (e.g. ImageNet),
% we count one prediction as correct when the estimated boxes have at least one IoU larger than 
% a larger IoU
% where the box prediction overlaps with at least
% one of the ground truth boxes with IoU ≥ δ
%%====================================================================================
% {%\small
% \begin{align}
%     MaxBoxAcc = max_{\tau}BoxAcc(\tau, \sigma)
%     \label{eq:MaxBoxAcc}
% \end{align}
% }%
% %\vspace{-6mm}
% {%\small
% \begin{align}
%     BoxAcc(\tau, \sigma) = \frac{1}{N}\sum_{n=0}^{N-1}1_{IoU(box(H, \tau), B^{n}) \geq \sigma}
%     \label{eq:BoxAcc}
% \end{align}
% }%
% $H$ is the generated localization map, $\tau$ is the threshold for $H$, $box(.)$ is the estimated bounding box given $H$ and $tau$, $B$ is the ground truth bounding box. $sigma$ is the threshold for the Intersection of Union ($IoU$). In the most works°\cite{}, $\sigma$ is set to 0.5. 
% \textit{MaxBoxAcc} measures the localization accuracy when $\sigma {=} 0.5$, while \textit{MaxBoxAccV2} measures the averaged \textit{MaxBoxAcc} when $\sigma {=} \{0.3, 0.5, 0.7\}$. 
% \textit{MaxBoxAcc} is equivalent to \textit{GT-known localization accuracy}~\cite{}.
% For multiple contours of the binarized localization map, following \cite{choe2020wsol}, we only consider the largest one.
%%====================================================================================
%
For OpenImages dataset, since we have access to the segmentation mask, we use the \textit{pixel averaged precision (PxAP)} proposed by \cite{choe2020wsol}. Similarly, \textit{PxAP} is also threshold-free.
Please refer to \cite{choe2020wsol} for more details.
%%====================================================================================
% {%\small
% \begin{align}
%     PxPrec(\tau) = \frac{|\{H^{n} \geq \tau\}\bigcap \{T^{n} = 1 \} |}{|\{H^{n} \geq \tau\}|}
%     \label{eq:pxprec}
% \end{align}
% }%
% %%\vspace{-1mm}
% {%\small
% \begin{align}
%     PxRec(\tau) = \frac{|\{H^{n} \geq \tau\}\bigcap \{T^{n} = 1 \} |}{|\{T^{n} = 1\}|}
%     \label{eq:pxrec}
% \end{align}
% }%
% $H$ and $T$ are generated localization heatmap and the segmentation mask, respectively.
% \textit{PxAP} is the area under curve of the pixel precision-recall curve.
%%====================================================================================

\subsection{Implementation Details}
% We implement our method via PyTorch~\cite{pytorch19}. 
We consider three different backbones: VGG16~\cite{VGG}, ResNet50~\cite{DBLP:journals/corr/HeZRS15} and the lightweight MobilenetV2~\cite{mobilenetv22018}.
Following \cite{choe2020wsol,zhang2020inter}, for ResNet50 and MobilenetV2 we increase the size of the last feature map by changing the stride of convolution layers from 2 to 1. 
By doing this, the model can still load the pretrained weights.
We set set size $S {=} 5$, $N {=} 12$ (i.e. batch size $= 60$) for all the experiments except those with the ResNet50 backbone, where $N {=} 10$ due to GPU memory limitations.
For $\tau$ we set 100 intervals between 0 and 1.

\subsection{Comparison with State-of-the-art Methods}
%%%=================================================
%%%=================================================
% In this experiment, we compare the proposed method w.r.t. several state-of-the-art methods on CUB and OpenImages datasets. Table~\ref{tab:QuantTable} shows quantitative results.
% % \textbf{[EXTEND]}
% The quantitative results from these methods except I2C are taken from \cite{choe2020wsol,wsol_eval_journal_submission} where the authors used the validation set to select the final models. 
% We implement I2C method and use their suggested hyper-parameters to train the models.
% %
% The results clearly show that our method outperforms the competing methods on both the CUB and OpenImages datasets except when ResNet50 is selected as backbone for the CUB dataset.
% % In this case, our method achieves competitive improvement w.r.t. HaS and ACoL for \textit{MaxBoxAcc} (0.3pp less) and \textit{MaxBoxAccV2} (1.2 pp less), respectively.
% In this case, our method is only lower by 0.1 pp, in \textit{MaxBoxAcc} w.r.t. HaS.
% % and 1.2 pp, in \textit{MaxBoxAccV2}, w.r.t. HaS and ACoL, respectively.
% It is expected that I2C works better when MobilenetV2 or ResNet50 are used as backbone, since the constraints proposed for I2C prevent the model from activating highly in background regions, which is a weakness that Mobilenet and ResNet suffer from.
%%%=================================================
%%%=================================================
In this part, we compare the proposed method w.r.t. several state-of-the-art methods on ImageNet, CUB and OpenImages datasets. Table~\ref{tab:QuantTable} shows quantitative results.
% \textbf{[EXTEND]}
The quantitative results from these methods except I2C are taken from \cite{choe2020wsol,wsol_eval_journal_submission} where the authors used the validation set to select the final models. 
We implement the I2C method and use their suggested hyper-parameters to train the models.
The results clearly show that our method outperforms the competing methods on all three datasets except when ResNet50 is selected as backbone for the CUB dataset.
% In this case, our method achieves competitive improvement w.r.t. HaS and ACoL for \textit{MaxBoxAcc} (0.3pp less) and \textit{MaxBoxAccV2} (1.2 pp less), respectively.
In this case, our method is only lower by 0.1 pp, in \textit{MaxBoxAcc} w.r.t. HaS.
% and 1.2 pp, in \textit{MaxBoxAccV2}, w.r.t. HaS and ACoL, respectively.
Interestingly, for ImageNet with a ResNet50 backbone, only our method outperforms CAM.
We believe it is due to the proposed $CRR$ which minimizes the activations in the background.
It is expected that I2C works better when MobilenetV2 or ResNet50 are used as backbone, since the constraints proposed by I2C prevent the model from activating highly in background regions, which is a weakness that Mobilenet and ResNet suffer from.
Please note 
% that we do not perform hyperparameter tunning. 
the performance
can be further improved when the optimal hyperparameters are found.
%%%=================================================
%%%=================================================

% Fig.~\ref{fig:qualityCUB} and Fig.~\ref{fig:qualityOpenimage} show a qualitative comparison w.r.t. CAM on the CUB and OpenImages datasets, respectively. It clearly shows that our method can enlarge $H$ when the original $H$ focuses on a small region and reduces it when it is activated highly in background regions.

Fig.~\ref{fig:qualityCUB} shows qualitative comparisons w.r.t. CAM on the ImageNet, CUB and OpenImages datasets. It clearly shows that our method can enlarge $H$ when it originally focuses on a small region and reduces it when it is highly-activated in the background.
In the fifth-column example from the ImageNet dataset generated by the VGG backbone, the effect of different optimal thresholds $\tau$ can be noticed. 
% It is interesting the fourth-column example from the ImageNet dataset generated by the VGG backbone, there it is noticeable the effect of different optimal thresholds $\tau$. 
%
The same effect can be seen in the Mobilenet-based localization map of the last example in the ImageNet dataset.
In addition, in some cases although the estimated bounding box of CAM has a large IoU with the ground truth, the object region has a stronger activation for our method (e.g. the last example of the ImageNet dataset). 

\begin{table*}[]
\centering
% %\small
\scalebox{0.72}{%
\begin{tabular}
{@{}lccccccc@{}}
\toprule
\multicolumn{1}{c}{\multirow{2}{*}{Method}} & \multirow{2}{*}{Backbone}& \multicolumn{2}{c}{ImageNet}& \multicolumn{2}{c}{CUB}& \multicolumn{2}{c}{OpenImages}\\ 
\cmidrule(l){3-6} 
\multicolumn{1}{c}{}                       &                                                                                                         & \begin{tabular}[c]{@{}c@{}}MaxBoxAcc (\%)\end{tabular}  &\begin{tabular}[c]{@{}c@{}}MaxBoxAccV2 (\%)\end{tabular} 
& \begin{tabular}[c]{@{}c@{}}MaxBoxAcc (\%)\end{tabular}  &\begin{tabular}[c]{@{}c@{}}MaxBoxAccV2 (\%)\end{tabular}&  ~~~~\begin{tabular}[c]{@{}c@{}}PxAP (\%)\end{tabular}   \\ \midrule
CAM & VGG16 &61.1~~~~~~~~~~&60.0~~~~~~~~~~ & 71.1~~~~~~~~~~ &  63.7~~~~~~~~~~  & 58.1~~~~~~~~   \\
HaS & VGG16 &~~~~~~~~~~0.7& ~~~~~~~~~~0.6 &  ~~~~~~~~~~5.2 &~~~~~~~~~~0 & ~~~~~~~~-1.2   \\ 
ACoL & VGG16 &~~~~~~~~~~-0.8&~~~~~~~~~~-2.6  & ~~~~~~~~~~1.2 & ~~~~~~~~~~-6.3 & ~~~~~~~~-3.4   \\ 
SPG & VGG16  &~~~~~~~~~~0.5& ~~~~~~~~~~-0.1& ~~~~~~~~~~-7.4 & ~~~~~~~~~~-7.4 & ~~~~~~~~-2.2   \\ 
ADL & VGG16  &~~~~~~~~~~-0.3& ~~~~~~~~~~-0.2& ~~~~~~~~~~4.6 & ~~~~~~~~~~2.6 & ~~~~~~~~0.2   \\ 
CutMix & VGG16 &~~~~~~~~~~1.0&~~~~~~~~~~-0.6  & ~~~~~~~~~~0.8 & ~~~~~~~~~~-1.4 & ~~~~~~~~0.1   \\ 
I2C & VGG16  &~~~~~~~~~~-&~~~~~~~~~~- & ~~~~~~~~~~-2.7 & ~~~~~~~~~~-3 & ~~~~~~~~-1   \\ 
Ours & VGG16 &~~~~~~~~~~\textit{\textbf{3.5}}&~~~~~~~~~~\textit{\textbf{2.2}} & ~~~~~~~~~~\textit{\textbf{12.8}} & ~~~~~~~~~~\textit{\textbf{6.5}} & ~~~~~~~~\textit{\textbf{1.9}}   \\ 
\midrule
CAM & ResNet50 &64.2~~~~~~~~~~&  63.7~~~~~~~~~~ & 73.2~~~~~~~~~~ & 63.0~~~~~~~~~~ & 58.0~~~~~~~~   \\
HaS & ResNet50 &~~~~~~~~~~-1& ~~~~~~~~~~-0.3 & ~~~~~~~~~~\textbf{4.9} & ~~~~~~~~~~1.7 & ~~~~~~~~0.2   \\ 
ACoL & ResNet50 &~~~~~~~~~~-2.5& ~~~~~~~~~~-1.4 &~~~~~~~~~~-0.5  & ~~~~~~~~~~3.5 & ~~~~~~~~-0.2   \\ 
SPG & ResNet50 &~~~~~~~~~~-0.7& ~~~~~~~~~~-0.4 & ~~~~~~~~~~-1.8 & ~~~~~~~~~~-2.6 & ~~~~~~~~-0.3   \\ 
ADL& ResNet50 &~~~~~~~~~~0& ~~~~~~~~~~0 & ~~~~~~~~~~0.3 & ~~~~~~~~~~-4.6 & ~~~~~~~~-3.7   \\ 
CutMix & ResNet50 &~~~~~~~~~~-0.3& ~~~~~~~~~~-0.4 & ~~~~~~~~~~-5.4 & ~~~~~~~~~~-0.2 & ~~~~~~~~-0.7   \\ 
I2C & ResNet50 &~~~~~~~~~~-&~~~~~~~~~~-  & ~~~~~~~~~~0.3 & ~~~~~~~~~~1.0 & ~~~~~~~~\textbf{2.9}   \\ 
Ours & ResNet50 &~~~~~~~~~~\textbf{\textit{2.5}}&~~~~~~~~~~\textit{\textbf{2.0}}  &~~~~~~~~~~\textit{4.8}  & ~~~~~~~~~~\textbf{\textit{4.3}} & ~~~~~~~~\textit{\textbf{2.9}}   \\ 
\midrule
CAM & MobilenetV2 &60.8~~~~~~~~~~& 59.5~~~~~~~~~~ & 65.3~~~~~~~~~~ & 58.1~~~~~~~~~~ & 54.9~~~~~~~~   \\
I2C& MobilenetV2  &~~~~~~~~~~-&~~~~~~~~~~- & ~~~~~~~~~~1.9 &~~~~~~~~~~1.5 & ~~~~~~~~3.3   \\ 
Ours & MobilenetV2 &~~~~~~~~~~\textbf{\textit{4.5}}&~~~~~~~~~~\textit{\textbf{3.8}}  & ~~~~~~~~~~\textit{\textbf{10.5}}  & ~~~~~~~~~~\textit{\textbf{6.9}} & ~~~~~~~~\textit{\textbf{4.4}}   \\ 
 \bottomrule
\end{tabular}%%\vspace{2mm}
}

% \caption{Quantitative comparison w.r.t. state-of-the-art. The numbers indicate the difference w.r.t. the baseline method CAM. The scores of CAM~\cite{zhou2016learning}, HaS~\cite{singh2017hide}, ACoL~\cite{zhang2018adversarial}, SPG~\cite{zhang2018self}, ADL~\cite{ADL2019}, CutMix~\cite{yun2019cutmix} are taken from \cite{choe2020wsol,wsol_eval_journal_submission}.
% Performance of I2C~\cite{zhang2020inter} was computed by ourselves.
% Due to limited computation resources we limit ourselves to report performance only on the CUB and OpenImages datasets
% %
% }
\caption{Quantitative comparison w.r.t. state-of-the-art. The numbers indicate the difference w.r.t. the baseline method CAM. The scores of CAM [Zhouet al., 2016], HaS [Singh and Lee, 2017], ACoL [Zhanget al., 2018a], SPG [Zhanget al., 2018b], ADL [Choe and Shim, 2019], CutMix [Yunet al., 2019] are taken from [Choeet al., 2020b; Choeet al., 2020a].
Performance of I2C was computed by ourselves.
Due to limited computation resources we limit ourselves to report performance only on the CUB and OpenImages datasets
}
\label{tab:QuantTable}
% \vspace{-3mm}
\end{table*}

\subsection{Analysis of our method}
\label{sec:analysis}
In this section we analyze several aspects of our method.
The experiments are conducted on the CUB dataset.

% We analyze several aspects of our method on the CUB dataset.

\subsubsection{Intensity change}
\label{sec:intensitychange}
% In order to make \textit{FRR} work, 
For \textit{FRR} to work,
$B$ should be sensitive to changes in intensity of its input.
To verify this assumption, we conduct an experiment where we add an intensity change as one of the data augmentations in stage I. In stage II we do not change anything.
By doing this, we force $B$ to become \textit{more} robust to intensity changes on its input
(Please note, we cannot make $B$ completely robust).
We use VGG16 as backbone. 
% Table~\ref{tab:intensitychange} shows the results. 
\tocheck{The performance drops 2.5pp and 1.9pp for MaxBoxAcc and MaxBoxAccV2, respectively.
This shows that indeed the performance drops when $B$ becomes more robust to intensity changes.}

%--------------------------------------------------------------------------
% \begin{table}
% \setlength{\tabcolsep}{4.7pt} % Default value: 6pt
% \centering
% % \footnotesize
% \begin{tabular*}{8cm}
% {@{\extracolsep{\fill}} l c c}
% \toprule 
% % Method & Try-on ROI (SSIM/LPSIS-VGG) & Take off (SSIM/LPSIS-VGG)  \\
% {Method}  & \textit{MaxBoxAcc} & \textit{MaxBoxAccV2}   \\ 

% \midrule[0.6pt]	
% w/ INTST change in DA & 81.4 &68.3 \\
% w/o INTST change in DA& 83.9&70.2\\

% \bottomrule[1pt]
% \end{tabular*}
% \caption{Performance comparison w.r.t. the robustness of intensity changes on the backbone $B$. \textbf{PUT THEM IN TEXT TO SAVE SPACE?}}
% \label{tab:intensitychange}
% %%\vspace{-4mm}
% \end{table}
%%--------------------------------------------------------------------------
%%--------------------------------------------------------------------------

\subsubsection{Background activation}
We have introduced in Sec.~\ref{sec:CRR} that $B$ should activate differently for different background regions for \textit{CRR} to work.
If the assumption holds, then $f {\in} \mathbb{R}^{K}$ ($f{=}B(I {\odot} H)$) from the same class should be distributed together and more centered in the K-dim feature space, compared with $f^o \in \mathbb{R}^{K}$ ($f^o {=} B(I)$).

\begin{figure}
\centering
\includegraphics[width=0.5\textwidth]{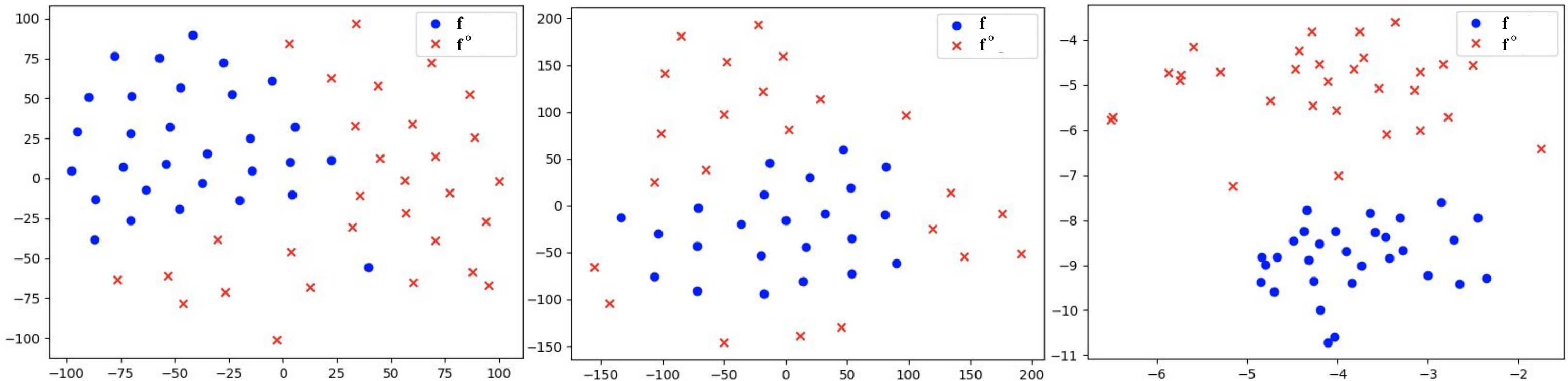}
\caption{t-SNE visualizations of $f$ and $f^{o}$ for class 64, 67 and 70 of CUB dataset.
The blue dots and red crosses refer to $f$ and $f^o$, respectively. Please zoom in for more details.}
\label{fig:tsne}
% \vspace{-4mm}
\end{figure}

To verify it, we first use t-SNE \cite{tsne} to visualize the features $f$ and $f^o$ from the same class, see Fig.~\ref{fig:tsne}.
Please note, we use the generated localization map $H$ to obtain $f$.
To quantitatively measure the distribution,
for each class, we first calculate the distance between each feature and the averaged feature of the class, then compute the standard deviation of the distances, and obtain the averaged standard deviation for all classes.
% Table~\ref{tab:classdistance} shows the statistic, 
For $f$ and $f^o$, the averaged standard deviations are $41.3 \pm 8.7$ and $51.4 \pm 10.3$, respectively, which suggests that $f$ is distributed more centered compared with $f^o$.  
The result proves our assumption, that indeed $B$ is activated differently for different backgrounds of the input image, and this is one of the reasons that \textit{CRR} works.

%%%%%%%%%-----------------------------------------------
%%%%%%%%%-----------------------------------------------
% %%\vspace{-4mm}

%%%%%%%%%-----------------------------------------------
%%%%%%%%%-----------------------------------------------

% %%--------------------------------------------------------------------------
% \begin{table}
% \setlength{\tabcolsep}{4.7pt} % Default value: 6pt
% \centering
% % \footnotesize
% \begin{tabular*}{8cm}
% {@{\extracolsep{\fill}} l c}
% \toprule 
% % Method & Try-on ROI (SSIM/LPSIS-VGG) & Take off (SSIM/LPSIS-VGG)  \\
% {Features}  & averaged standard deviation \\ 
% \midrule[0.6pt]	
% $f = B(I \odot H)$ & 41.3 $\pm$ 8.7 \\
% $f^o = B (I)$ & 51.4 $\pm$ 10.3\\
% \bottomrule[1pt]
% \end{tabular*}
% \caption{Averaged standard deviation (over 200 classes) of the distance between each feature to its corresponding mean feature. Lower number indicates more centered distribution in the feature space.}
% \label{tab:classdistance}
% \end{table}
% %%--------------------------------------------------------------------------
% %%--------------------------------------------------------------------------

% \subsubsection{Location to apply $H$}
\subsubsection{Masking inputs vs. masking features}
In order to get the
% the representation of the region focused on the object,
localized-object representation $f$,
we suggest applying element-wise product between $H$ and the input image $I$ firstly and then send it to the same backbone $B$.
It can be argued that $H$ can be applied on the extracted feature map (noted as $f' {\in} \mathbb{R}^{N{\times} C {\times} H {\times} W}$) from stage I directly, which can avoid the re-computation of the feature.
However, this may produce some problems.
% for the quality of the feature maps.

On the one hand, 
$f'$ not only has spatial information but also has $C$ channels (around 1024 or 2048 for our backbones), and  different channels can represent different concepts of the image \cite{kaili:visualExpInter}.
On the other hand, the localization map $H {\in} \mathbb{R}^{H {\times} W}$ only contains spatial information, no channel-wise information.
\tocheck{Information can be lost if $H$ is directly applied on $f'$ since the activations on the same spatial location but different channels will be encouraged/suppressed in the same scale.}
To verify our analysis, we conduct an experiment where we apply $H$ directly on $f'$.
Table~\ref{tab:locationH} suggests that our analysis is correct. Especially, for VGG16
% (i.e. the models that $FRR$ dominate, we will discuss it in Sec.~\ref{sec:ablationstudy}),
the performance is even worse than CAM. 

%%--------------------------------------------------------------------------
% \begin{table}
% \setlength{\tabcolsep}{4.7pt} % Default value: 6pt
% \centering
% % \footnotesize
% \begin{tabular*}{8cm}
% {@{\extracolsep{\fill}} l c c}
% \toprule 
% % Method & Try-on ROI (SSIM/LPSIS-VGG) & Take off (SSIM/LPSIS-VGG)  \\
% {Method}  & \textit{MaxBoxAcc} (\%) & \textit{MaxBoxAccV2}(\%)   \\ 

% \midrule[0.6pt]	
% % VGG16-CAM & 71.1 &63.7\\
% Ours-VGG16 & 83.9& 70.2\\
% Apply $H$ on $f'$ & -19.1 &-12.5 \\
% \midrule[0.6pt]	
% Ours-Mobilenet & 75.8& 65.0\\
% Apply $H$ on $f'$ & -4.1 & -3.0 \\
% \bottomrule[1pt]
% \end{tabular*}
% \caption{Location to apply $H$.}
% \label{tab:locationH}
% %%\vspace{-4mm}
% \end{table}
%%--------------------------------------------------------------------------
%%--------------------------------------------------------------------------
%%--------------------------------------------------------------------------
%%--------------------------------------------------------------------------
\begin{figure*}
\begin{minipage}[b]{0.35\textwidth}
\centering
\resizebox{1\columnwidth}{!}{
\setlength{\tabcolsep}{3pt}
\renewcommand{\arraystretch}{1.55}
\begin{tabular*}{8cm}
{@{\extracolsep{\fill}} l c c}
\toprule 
% Method & Try-on ROI (SSIM/LPSIS-VGG) & Take off (SSIM/LPSIS-VGG)  \\
{Method}  & \textit{MaxBoxAcc} (\%) & \textit{MaxBoxAccV2}(\%)   \\ 

\midrule[0.6pt]	
% VGG16-CAM & 71.1 &63.7\\
VGG-Ours & 83.9& 70.2\\
Apply $H$ on $f'$ & -19.1 &-12.5 \\
\midrule[0.6pt]	
MobilenetV2-Ours & 75.8& 65.0\\
Apply $H$ on $f'$ & -4.1 & -3.0 \\
\bottomrule[1pt]
\end{tabular*}
}
\captionof{table}{
Masking inputs vs. masking features
% Location to apply $H$.
}
\label{tab:locationH}
\end{minipage}
\hskip 1pt
\begin{minipage}[b]{0.33\textwidth}
\centering
\resizebox{1\columnwidth}{!}{
\setlength{\tabcolsep}{1pt}
\renewcommand{\arraystretch}{1.4}
\begin{tabular*}{8cm}
{@{\extracolsep{\fill}} l c c}
\toprule 
% Method & Try-on ROI (SSIM/LPSIS-VGG) & Take off (SSIM/LPSIS-VGG)  \\
{Set size $S$}  & \textit{MaxBoxAcc(\%)} & \textit{MaxBoxAccV2(\%)}   \\ 

\midrule[0.6pt]	
$S{=}5$& 75.8 &65.0 \\
$S {=}4$& 74.7 & 64.6\\
$S{=}3$&74.4 & 64.4\\
$S{=}2$&73.9 & 63.9\\
\midrule[0.6pt]	
CAM & 65.3 & 58.1\\
\bottomrule[1pt]
\end{tabular*}
}
\captionof{table}{Effect of the set size $S$.}
\label{tab:setSizeEffect}
\end{minipage}
\hskip 1pt
\begin{minipage}[b]{0.3\textwidth}
\centering
\resizebox{1\columnwidth}{!}{
\begin{tabular*}{7cm}
{@{\extracolsep{\fill}} l c}
\toprule 
% Method & Try-on ROI (SSIM/LPSIS-VGG) & Take off (SSIM/LPSIS-VGG)  \\
{Method}  & \textit{Top 1 Loc. / Top 5 Loc.} \\ 
\midrule[0.6pt]	
VGG-ACoL & 45.9~/~~-~~~~ \\
VGG-ADL & 52.3~/~~-~~~~ \\
VGG-CCAM & 50.1~/~63.8\\
VGG-Ours & \textit{\textbf{66.0}}~/~\textit{\textbf{83.9}} \\
% VGG-ADL & 52.4 /~~~ ~~-\\
% \midrule[0.6pt]	
% ResNet-Ours & 60.6 / 76.6 \\
\midrule[0.6pt]	
MobilenetV1-HaS & 44.7 / ~~-~~~~\\
MobilenetV1-ADL & 47.7 / ~~-~~~~\\
MobilenetV1-Ours & \textit{\textbf{54.8}} /~\textit{\textbf{69.4}}  \\
% MobilenetV2-Ours & 60.0 / 75.0 \\
\bottomrule[1pt]
\end{tabular*}
}
\captionof{table}{Top-1 and Top-5 localization rate.}
\label{tab:top15}
\end{minipage}
\end{figure*}
%%--------------------------------------------------------------------------
%%--------------------------------------------------------------------------

\subsection{Ablation study}
\label{sec:ablationstudy}
\subsubsection{Effects of \textit{CRR} and \textit{FRR} for different backbones ($B$)}
We analyze the effect of \textit{CRR} and \textit{FRR} on different backbones.
In Sec.~\ref{sec:problemstatement} we mentioned the issue of the localization map $H$ generated by different architectures with Fig.~\ref{fig:shortage} showing some examples.
To verify that the failure cases are consistent across the whole dataset,
% we firstly analyse how the generated localization map is distributed on the image.
we analyse how large is the background covered by the estimated bounding boxes 
inferred 
% from the localization map 
from the VGG16 and MobilenetV2 backbones.

For each test image, we compute the proportion of pixels in the predicted bounding box that cover background regions.
The higher the proportion, the more background is covered by the estimated bounding box.

% If there is no intersection, then $S_{BG esti} = $
% we obtain the localization map and estimate the bounding box.
% We compute the area of the intersection of estimated and GT bounding box and the background area is 
% Afterwards, we calculate the ratio of the intersection area and input image size.  
%
% If the estimated box is inside of the GT box, the background area is 0.
For MobilenetV2, roughly half of the images have a proportion above 0.1,
while that happens only for roughly a quarter of the images for VGG16.
This proves that indeed the localization map generated by MobilenetV2 activates more in the background region.
Please refer to the supplementary material for more details.
%%%%%%%%%-----------------------------------------------
%%%%%%%%%-----------------------------------------------
% \begin{figure}
% \centering
% \includegraphics[width=0.4\textwidth]{./imgs/ratio_vgg_mobile_v2.pdf}
% \caption{Activation distribution of the generated localization map. small conclusion in caption, }
% \label{fig:ratio}
% %%\vspace{-4mm}
% \end{figure}
% % %%\vspace{-4mm}

%%%%%%%%%-----------------------------------------------
%%%%%%%%%-----------------------------------------------

In practice, intuitively, if the localization map $H$ always localizes the most discriminative region of the object, \textit{FRR} should play a more important role in the training process. On the contrary, \textit{CRR} should be more essential if $H$ is relatively highly activated in background regions.
% We conduct two experiments to verify this: we disable the \textit{FRR} and \textit{CRR} for VGG16 and Mobilenet, respectively. This depicts the scenario described above.
% Table~\ref{tab:crrfrreffect} clearly shows our intuition is correct, the performance drops drastically without using \textit{FRR} for VGG16 and \textit{CRR} for Mobilenet.
To verify the influence of \textit{CRR} and \textit{FRR},
we 
%select VGG16 and MobilenetV2 as backbones and 
gradually increase/decrease the weight of one regularization with the other one fixed.

\begin{figure}
\centering
\includegraphics[width=0.5\textwidth]{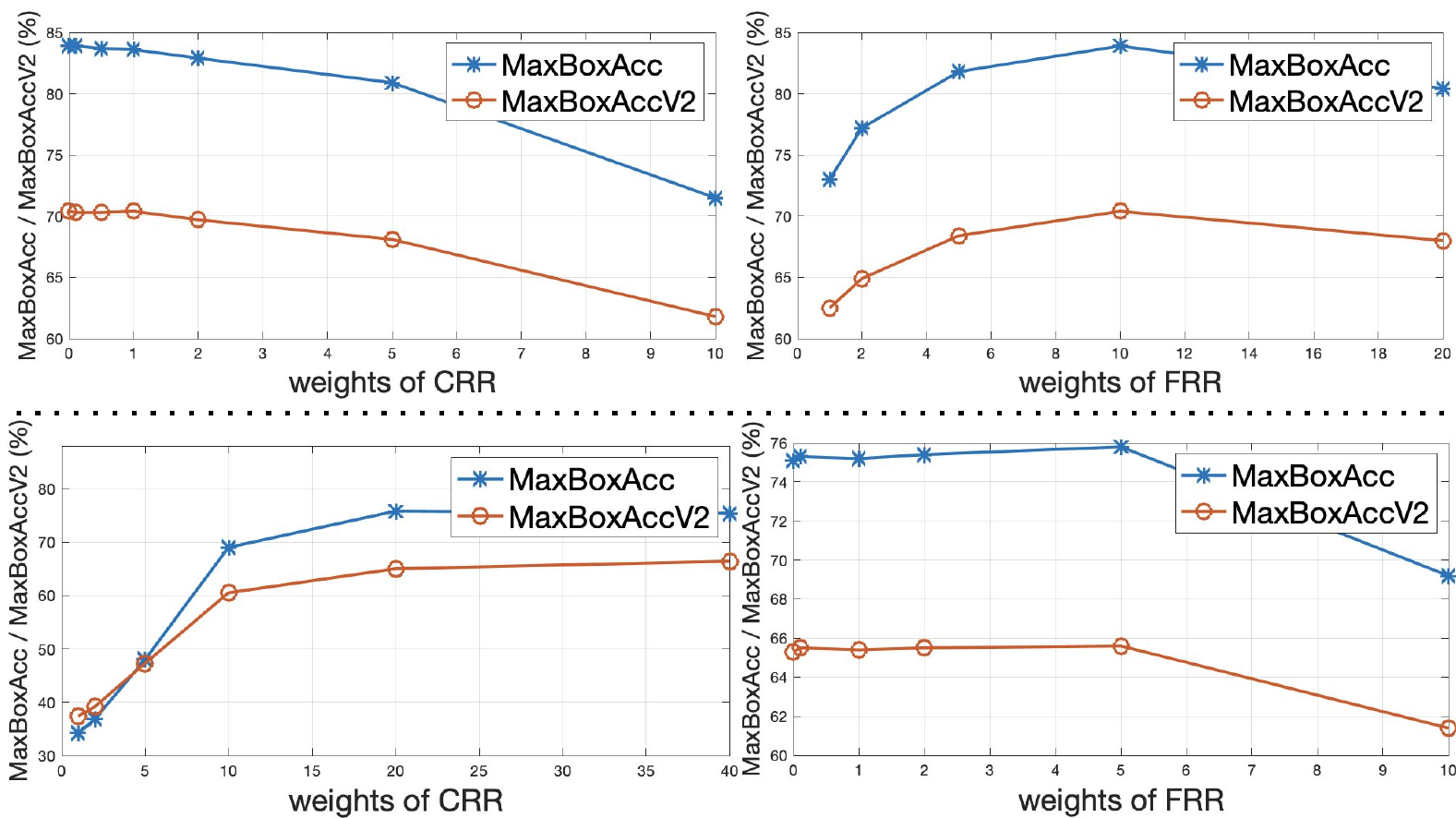}
\caption{Ablation study w.r.t. \textit{CRR} and \textit{FRR} for the VGG16 (top) and MobilenetV2 (bottom) backbones, respectively. For each plot, We fix one regularization and ablate the other.
%
% The plots above the dashed line are the experiments with VGG16 backbone while the blow two are with MobilenetV2 backbone. For each plot, We ablate one regularization with another one fixed.
}
\label{fig:crrfrreffect}
% \vspace{-4mm}
\end{figure}

Fig.~\ref{fig:crrfrreffect} shows the performance curve.
For VGG16, performance decreases gradually as the weight for \textit{CRR} increases. The opposite occurs with \textit{FRR}. The performance increases and reach the peak when the weight of \textit{FRR} is 10, afterwards the performance decreases slightly (weight{=}20).
On the contrary, increasing the weight of \textit{CRR} boosts the performance for MobilenetV2.
The performance decreases slightly when the weight of \textit{FRR} is 0 while it drops dramatically when a larger weight is applied.
%
% These trends are expected since VGG16 normally fails on localizing the most discriminative region rather than the whole object
% and what MobilenetV2 fails is opposite.
These trends are expected since VGG16 focuses on small discriminative regions rather than the whole object while MobilenetV2 activates frequently on the background. 
In addition, the experiment with MobilenetV2 shows that \textit{FRR} can compensate the performance when a large \textit{CRR} is needed and applied.

% To verify this 

% %%--------------------------------------------------------------------------
% \begin{table}
% \setlength{\tabcolsep}{4.7pt} % Default value: 6pt
% \centering
% % \footnotesize
% \begin{tabular*}{8cm}
% {@{\extracolsep{\fill}} l c c}
% \toprule 
% % Method & Try-on ROI (SSIM/LPSIS-VGG) & Take off (SSIM/LPSIS-VGG)  \\
% {Method}  & \textit{MaxBoxAcc} (\%) & \textit{MaxBoxAccV2}(\%)   \\ 

% \midrule[0.6pt]	
% % VGG16-CAM & 71.1 &63.7\\
% Oours-VGG16 & 83.9& 70.3\\
% w/o \textit{FRR} & -11.8 &-11.4 \\
% w/o \textit{CRR} & 0 &0 \\
% \midrule[0.6pt]	
% % Mobilenet-CAM & 68.6& 59.9\\
% Ours-MobileNet & 75.2&65.4\\
% w/o \textit{FRR} & 0 & 0 \\
% w/o \textit{CRR} & -9.7 &-6.8 \\
% \midrule[0.6pt]	
% % Mobilenet-CAM & 68.6& 59.9\\
% Ours-ResNet50 & 76.9&66.2\\
% w/o \textit{FRR} & -0.6 & -0.3 \\
% w/o \textit{CRR} & -10.9 &-7.2 \\
% \bottomrule[1pt]
% \end{tabular*}
% \caption{Ablation study on the effect of \textit{CRR} and \textit{FRR}.
% \textbf{PLACE HOLDER:REPLACE BY FIGURE}}
% \label{tab:crrfrreffect}
% \end{table}
% %%--------------------------------------------------------------------------
% %%--------------------------------------------------------------------------
%%%%%%%%%-----------------------------------------------
%%%%%%%%%-----------------------------------------------
% \vspace{-2mm}

%%%%%%%%%-----------------------------------------------
%%%%%%%%%-----------------------------------------------

\subsubsection{Effect of the set size ($S$)}
Here we discuss how the set size $S$ influences the performance of \textit{CRR}.
\textit{CRR} can benefit from using a relatively large $S$ because more representations can be grouped together simultaneously.
We conduct series of experiments, where we decrease $S$ from $5$ to $2$ gradually. 
We use Mobilenet as backbone and keep the batch size the same (60 images).
The results in Table~\ref{tab:setSizeEffect} indicate that for a larger set size $S$, \textit{FRR} indeed works better.
%%--------------------------------------------------------------------------
% \begin{table}
% \setlength{\tabcolsep}{4.7pt} % Default value: 6pt
% \centering
% % \footnotesize
% \begin{tabular*}{8cm}
% {@{\extracolsep{\fill}} l c c}
% \toprule 
% % Method & Try-on ROI (SSIM/LPSIS-VGG) & Take off (SSIM/LPSIS-VGG)  \\
% {Set size $S$}  & \textit{MaxBoxAcc} & \textit{MaxBoxAccV2}   \\ 

% \midrule[0.6pt]	
% $S{=}5$& 75.8 &65.0 \\
% $S {=}4$& 74.7 & 64.6\\
% $S{=}3$&74.4 & 64.4\\
% $S{=}2$&73.9 & 63.9\\
% \midrule[0.6pt]	
% CAM & 65.3 & 58.1\\

% \bottomrule[1pt]
% \end{tabular*}
% \caption{Effect of the set size $S$.}
% \label{tab:setSizeEffect}
% \end{table}

\subsection{Classification v.s. localization}
The classification task sometimes rely on information from the background \cite{kaili:visualExpInter}, while the localization task only focuses on the foreground object. Therefore, a good localization model is not necessarily a good classification model.
The evaluation metric \textit{top-1/5 Loc.} takes into account both localization and classification accuracy of a given localization model,therefore, it is not able to accurately measure the localization performance
% which cannot evaluate a localization model properly
\cite{choe2020wsol}.
\textit{Top-1/5 Loc.} can be low because of the classification accuracy even if the localization accuracy is good.
% Therefore, following \cite{choe2020wsol}, we use \textit{MaxBoxAcc(V2)} to evaluate the localization model.

In order to compute \textit{Top-1/5 Loc.} fairly, we propose a simple path: train a separate classification model to provide the predicted class for the object localization model.
In practice, we use ResNet50 to train the classifier, whose classification accuracy on CUB dataset is 77.3\%. Table~\ref{tab:top15} shows the \textit{Top-1/5 Loc.} results. 
In order to compare with the competitive methods, here we use MobilenetV1 as backbone.
The numbers of the competitive methods are taken from \cite{ADL2019,CCAM_20}.
%%--------------------------------------------------------------------------
% \begin{table}
% \setlength{\tabcolsep}{4.7pt} % Default value: 6pt
% \centering
% % \footnotesize
% \scalebox{0.9}{%
% \begin{tabular*}{8cm}
% {@{\extracolsep{\fill}} l c}
% \toprule 
% % Method & Try-on ROI (SSIM/LPSIS-VGG) & Take off (SSIM/LPSIS-VGG)  \\
% {Method}  & \textit{Top 1 Loc. / Top 5 Loc.} \\ 
% \midrule[0.6pt]	
% VGG-ACoL & 45.9~/~~-~~~~ \\
% VGG-ADL & 52.3~/~~-~~~~ \\
% VGG-CCAM & 50.1~/~63.8\\
% VGG-Ours & \textit{\textbf{66.0}}~/~\textit{\textbf{83.9}} \\
% % VGG-ADL & 52.4 /~~~ ~~-\\
% % \midrule[0.6pt]	
% % ResNet-Ours & 60.6 / 76.6 \\
% \midrule[0.6pt]	
% MobilenetV1-HaS & 44.7 / ~~-~~~~\\
% MobilenetV1-ADL & 47.7 / ~~-~~~~\\
% MobilenetV1-Ours & \textit{\textbf{54.8}} /~\textit{\textbf{69.4}}  \\
% % MobilenetV2-Ours & 60.0 / 75.0 \\
% \bottomrule[1pt]
% \end{tabular*}
% }
% \caption{Top 1 and Top 5 localization rate.}
% \label{tab:top15}
% %%\vspace{-4mm}
% \end{table}
%%--------------------------------------------------------------------------
%%--------------------------------------------------------------------------
%%\vspace{-0mm}
% We provide the quantitative results in the supplementary material.
\subsection{Failure cases}
\begin{figure}
\centering
\includegraphics[width=0.5\textwidth]{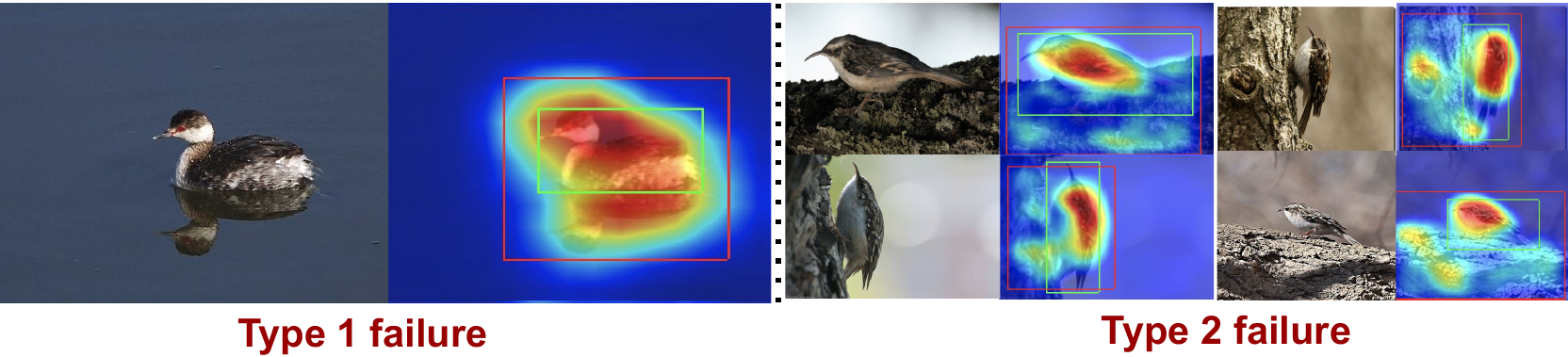}
\caption{Failure cases of the proposed method.
% The left and right examples show the first and second type, respectively.
}
\label{fig:failurecase}
% \vspace{-4mm}
\end{figure}

We analyze some failure cases in this section.
The first type is caused by nature, which is unavoidable, like reflections of water.
The cause of the second type is related to our first assumption for \textit{CRR}, that different images from a class have very similar background. For example, brown creeper always appear with trunks in the image.
Fig.~\ref{fig:failurecase} shows some examples.
% Please see the examples in the supplementary material.
%%%%%%%%%-----------------------------------------------
%%%%%%%%%-----------------------------------------------
% \vspace{-6mm}

%%%%%%%%%-----------------------------------------------
%%%%%%%%%-----------------------------------------------

\section{Conclusion}
We propose two representation regularizations, \textit{Common Region Regularization} and \textit{Full Region Regularization}, to overcome the weaknesses of weakly supervised object localization methods based on CAM.
Our method relies only on a standard classification model; no extra network is needed.
% Through extensive experiments and analysis, we show
% that our method is effective, surpassing existing methods by a significant margin 
% and we reveal the insight that why the proposed method works.
%
Through extensive experiments and analysis, we discuss relevant aspects of our method and show that it is capable of
surpassing the state-of-the-art by a significant margin.
\newpage
% \newpage
\bibliographystyle{named}
\bibliography{ijcai21}
\end{document}